\PassOptionsToPackage{unicode}{hyperref}
\PassOptionsToPackage{hyphens}{url}
\PassOptionsToPackage{dvipsnames,svgnames,x11names}{xcolor}
\documentclass[
  number,
  preprint,
  3p]{elsarticle}

\usepackage{amsmath,amssymb}
\usepackage{iftex}
\ifPDFTeX
  \usepackage[T1]{fontenc}
  \usepackage[utf8]{inputenc}
  \usepackage{textcomp} 
\else 
  \usepackage{unicode-math}
  \defaultfontfeatures{Scale=MatchLowercase}
  \defaultfontfeatures[\rmfamily]{Ligatures=TeX,Scale=1}
\fi
\usepackage{lmodern}
\ifPDFTeX\else  
\fi
\IfFileExists{upquote.sty}{\usepackage{upquote}}{}
\IfFileExists{microtype.sty}{
  \usepackage[]{microtype}
  \UseMicrotypeSet[protrusion]{basicmath} 
}{}
\makeatletter
\@ifundefined{KOMAClassName}{
  \IfFileExists{parskip.sty}{%
    \usepackage{parskip}
  }{
    \setlength{\parindent}{0pt}
    \setlength{\parskip}{6pt plus 2pt minus 1pt}}
}{
  \KOMAoptions{parskip=half}}
\makeatother
\usepackage{xcolor}
\makeatletter
\ifx\paragraph\undefined\else
  \let\oldparagraph\paragraph
  \renewcommand{\paragraph}{
    \@ifstar
      \xxxParagraphStar
      \xxxParagraphNoStar
  }
  \newcommand{\xxxParagraphStar}[1]{\oldparagraph*{#1}\mbox{}}
  \newcommand{\xxxParagraphNoStar}[1]{\oldparagraph{#1}\mbox{}}
\fi
\ifx\subparagraph\undefined\else
  \let\oldsubparagraph\subparagraph
  \renewcommand{\subparagraph}{
    \@ifstar
      \xxxSubParagraphStar
      \xxxSubParagraphNoStar
  }
  \newcommand{\xxxSubParagraphStar}[1]{\oldsubparagraph*{#1}\mbox{}}
  \newcommand{\xxxSubParagraphNoStar}[1]{\oldsubparagraph{#1}\mbox{}}
\fi
\makeatother

\providecommand{\tightlist}{%
  \setlength{\itemsep}{0pt}\setlength{\parskip}{0pt}}\usepackage{longtable,booktabs,array}
\usepackage{calc} 
\usepackage{etoolbox}
\makeatletter
\patchcmd\longtable{\par}{\if@noskipsec\mbox{}\fi\par}{}{}
\makeatother
\IfFileExists{footnotehyper.sty}{\usepackage{footnotehyper}}{\usepackage{footnote}}
\makesavenoteenv{longtable}
\usepackage{graphicx}
\makeatletter
\def\maxwidth{\ifdim\Gin@nat@width>\linewidth\linewidth\else\Gin@nat@width\fi}
\def\maxheight{\ifdim\Gin@nat@height>\textheight\textheight\else\Gin@nat@height\fi}
\makeatother
\setkeys{Gin}{width=\maxwidth,height=\maxheight,keepaspectratio}
\makeatletter
\def\fps@figure{htbp}
\makeatother

\makeatletter
\@ifpackageloaded{caption}{}{\usepackage{caption}}
\AtBeginDocument{%
\ifdefined\contentsname
  \renewcommand*\contentsname{Table of contents}
\else
  \newcommand\contentsname{Table of contents}
\fi
\ifdefined\listfigurename
  \renewcommand*\listfigurename{List of Figures}
\else
  \newcommand\listfigurename{List of Figures}
\fi
\ifdefined\listtablename
  \renewcommand*\listtablename{List of Tables}
\else
  \newcommand\listtablename{List of Tables}
\fi
\ifdefined\figurename
  \renewcommand*\figurename{Figure}
\else
  \newcommand\figurename{Figure}
\fi
\ifdefined\tablename
  \renewcommand*\tablename{Table}
\else
  \newcommand\tablename{Table}
\fi
}
\@ifpackageloaded{float}{}{\usepackage{float}}
\floatstyle{ruled}
\@ifundefined{c@chapter}{\newfloat{codelisting}{h}{lop}}{\newfloat{codelisting}{h}{lop}[chapter]}
\floatname{codelisting}{Listing}

\makeatother
\makeatletter
\@ifpackageloaded{caption}{}{\usepackage{caption}}
\@ifpackageloaded{subcaption}{}{\usepackage{subcaption}}
\makeatother
\journal{Journal Name}

\ifLuaTeX
  \usepackage{selnolig}  
\fi
\usepackage[]{natbib}
\usepackage{bookmark}

\IfFileExists{xurl.sty}{\usepackage{xurl}}{} 
\hypersetup{
  pdftitle={Enhancing Feature Selection and Interpretability in AI Regression Tasks Through Feature Attribution},
  pdfauthor={Alexander Hinterleitner; Thomas Bartz-Beielstein; Richard Schulz; Sebastian Spengler; Thomas Winter; Christoph Leitenmeier},
  pdfkeywords={XAI, Explainable artificial intelligence, Feature
attribution, Feature selection, Deep neural network, Regression},
  colorlinks=true,
  linkcolor={blue},
  filecolor={Maroon},
  citecolor={Blue},
  urlcolor={Blue},
  pdfcreator={LaTeX via pandoc}}

\begin{document}

\begin{frontmatter}
\title{Enhancing Feature Selection and Interpretability in AI Regression
Tasks Through Feature Attribution \\\large{A Case Study of Predicting
Blade Vibrations in Turbo Machinery} }
\author[1]{Alexander Hinterleitner%
\corref{cor1}%
}
 \ead{alexander.hinterleitner@th-koeln.de} 
\author[1]{Thomas Bartz-Beielstein%
}
 \ead{thomas.bartz-beielstein@th-koeln.de} 
\author[1]{Richard Schulz%
}
 \ead{richard.schulz@th-koeln.de} 
\author[2]{Sebastian Spengler%
}
 \ead{sebastian.spengler@man-es.com} 
\author[2]{Thomas Winter%
}
 \ead{thomas.winter@man-es.com} 
\author[2]{Christoph Leitenmeier%
}
 \ead{christoph.leitenmeier@man-es.com} 

\affiliation[1]{organization={TH Koeln, Institute for Data Science,
Engineering \& Analytics},addressline={Steinmueller Allee
1},city={Gummersbach},postcode={51643},postcodesep={}}
\affiliation[2]{organization={MAN Energy
Solutions},addressline={Stadtbachstraße
1},city={Augsburg},postcode={86224},postcodesep={}}

\cortext[cor1]{Corresponding author}

\begin{abstract}
Research in Explainable Artificial Intelligence (XAI) is increasing,
aiming to make deep learning models more transparent. Most XAI methods
focus on justifying the decisions made by Artificial Intelligence (AI)
systems in security-relevant applications. However, relatively little
attention has been given to using these methods to improve the
performance and robustness of deep learning algorithms. Additionally,
much of the existing XAI work primarily addresses classification
problems. In this study, we investigate the potential of feature
attribution methods to filter out uninformative features in input data
for regression problems, thereby improving the accuracy and stability of
predictions. We introduce a feature selection pipeline that combines
Integrated Gradients with k-means clustering to select an optimal set of
variables from the initial data space. To validate the effectiveness of
this approach, we apply it to a real-world industrial problem - blade
vibration analysis in the development process of turbo machinery.
\end{abstract}

\begin{keyword}
    XAI \sep Explainable artificial intelligence \sep Feature
attribution \sep Feature selection \sep Deep neural network \sep 
    Regression
\end{keyword}
\end{frontmatter}

\section{Introduction}\label{introduction}

Neural networks have proven to be universal function approximators,
especially for high dimensional and complex data sets
\citep{lecun2015deep}\citep{yang2013investigation}. However, despite
their powerful capabilities, they often require substantial
computational resources, and their predictions lack interpretability due
to their black-box nature. While neural networks can learn complex
representations, feature selection may enhance efficiency and
performance \citep{verikas2002feature}\citep{navot2005feature}. In the
context of neural networks, classical approaches such as wrapper methods
are often computationally prohibitive due to the necessity of multiple
training and evaluation cycles. Filter methods select features by
looking at their statistical properties (e.g., correlation, mutual
information) with respect to the target variable. These methods do not
involve the model itself and tend to capture only linear or simple
relationships between features and the target. Since neural networks are
known for their ability to model complex, non-linear patterns in data,
filter methods are often not sufficient in such cases, as they may not
be able to detect important feature interactions that only become
apparent when modeling more complex structures.

In recent years, various explainable artificial intelligence (XAI)
approaches, particularly feature attribution methods, have been
developed to make the decision-making processes of neural networks more
transparent and to understand which features influence their predictions
\citep{selvaraju2017grad}\citep{lundberg2017unified}\citep{ribeiro2016should}\citep{shrikumar2017learning}\citep{sundararajan2017axiomatic}\citep{bach2015pixel}.
While the majority of these methods have focused on classification
problems, there are a few notable efforts addressing feature attribution
in regression problems \citep{Mamalakis2023}\citep{Letzgus2022}.
Additionally, most work in this field has focused on explaining the
decisions of models for legal or ethical purposes rather than using it
for Machine Learning Engineering. Biecek and Samek (2024)
\citep{biecek2024explain} differentiate between these two scopes of
explanations by categorizing them into BLUE (responsi\textbf{B}le
models, \textbf{L}egal issues, tr\textbf{U}st in predictions,
\textbf{E}thical issues) and RED (\textbf{R}esearch on data,
\textbf{E}xplore models, \textbf{D}ebug models) XAI.

Our research focuses on investigating the potential of a popular group
of XAI methods, known as feature attribution methods, for feature
selection to filter out uninformative data and enhance neural network
performance in a real-world regression problem related to blade
vibration analysis in turbomachinery. We argue that models can identify
patterns in the data that may be unknown to domain experts. Therefore,
using explanations of a specific network can be beneficial in
determining the optimal parameter set through a reverse engineering
process. We employ Integrated Gradients (IG), a gradient-based and
model-agnostic approach that has been previously utilized in regression
contexts \citep{Mamalakis2023}\citep{Letzgus2022}. We propose a
data-driven process that combines the aggregation of local attribution
values for a global perspective with the k-Means clustering algorithm to
ensures the selection of only the most relevant features for the neural
network's decision-making process. While most work in this field employs
out-of-the-box neural network architectures that are analyzed using
feature attribution methods, we use surrogate-based hyperparameter
optimization to find the best hyperparameter setting for our model. This
is crucial because we argue that the information provided by the
gradient-based attribution method is only as valuable as the prediction
quality of our model. Silva et al.~investigated this relationship
between model performance and feature attribution
\citep{silva2024exploring}. To validate IG, we first conduct experiments
using generated, transparent dummy data before applying it to real-world
problems. We also compare our method with established base line feature
selection approaches and KernelShap, an approximation of one of the most
famous attribution methods known as Shapley values.

\section{Related Work}\label{related-work}

\subsection{Classical Feature Selection
Methods}\label{classical-feature-selection-methods}

Feature selection is an important part of the preprocessing stage in
machine learning applications. Especially for high-dimensional problems,
the reduction of input features can speed up the training of the
algorithm and make the problem more interpretable. The prediction
quality of the model can also be improved by eliminating disturbance
variables. Classic feature selection methods can be categorized as
filters, wrappers, and embedded methods
\citep{jovic2015review}\citep{li2017feature}\citep{cai2018feature}.

Filter methods for feature selection are computationally efficient
techniques that evaluate the relevance of features independently of any
specific learning algorith. These methods typically employ statistical
measures to score and rank features based on their correlation with the
target variable. Common filter approaches include the Pearson
correlation coefficient \citep{liu2020daily}, mutual information
\citep{vergara2014review}, Chi-squared test
\citep{thaseen2017intrusion}, and information gain
\citep{azhagusundari2013feature}. The key advantage of filter methods is
their speed and scalability, making them suitable for high-dimensional
datasets. However, they usually do not consider feature interactions or
complex non-linear correlations. These methods are often used as
preprocessing steps or baselines for more sophisticated methods
\citep{sanchez2007filter}\citep{cherrington2019feature}.

The second category of feature selection algorithms is wrapper methods.
These methods are model-dependent and use the model's performance as a
criterion to evaluate feature subsets, effectively ``wrapping'' the
selection process around the model itself \citep{kohavi1997wrappers}.
One common approach within this category is forward selection, where the
process begins with an empty feature set and iteratively adds features
that maximize model performance. At each iteration, the model is trained
on the current feature subset along with each remaining feature
individually, and the feature that yields the highest performance
improvement is permanently added to the subset. This continues until no
significant improvement is observed or a predetermined number of
features is reached. Wrapper approaches tend to be more accurate than
filter methods but are generally higher in their computational costs
\citep{ang2015supervised}, especially for high-dimensional problems and
Deep Learning applications.

The third category of methods is defined as embedded methods, which
integrate feature selection directly into the model training process
\citep{wang2015embedded}\citep{lal2006embedded}. Unlike filter methods
that do not incorporate learning, and wrapper methods that evaluate
feature subsets using a learning machine without considering the
specific structure of the classification or regression model, embedded
methods combine the learning and feature selection processes. Examples
of embedded methods include tree algorithms
\citep{quinlan1986induction}\citep{breiman2001random} or linear models
and related algorithms like Lasso or Ridge regression
\citep{tibshirani1996regression}\citep{hoerl1970ridge}.

\subsection{Feature Attribution}\label{feature-attribution}

Feature attribution methods have become essential tools for interpreting
complex machine learning models. These methods aim to quantify the
importance of input features on a model's predictions, thereby
increasing model transparency and aiding feature selection. In this
section, we review state-of-the-art feature attribution methods,
highlighting both model-agnostic and model-specific approaches.

Introduced by Lundberg and Lee (2017) \citep{lundberg2017unified}, SHAP
(SHapley Additive exPlanations) consolidates several prior attribution
methods into a unified framework based on Shapley values from
cooperative game theory. This approach provides a consistent and
theoretically sound method for feature attribution, maintaining
properties such as local accuracy and consistency. Its versatility
allows it to effectively analyze feature importance across various
machine learning models. However, it is important to note that SHAP's
computational demands can sometimes be prohibitive, especially with
large datasets and complex models, as it requires evaluating numerous
feature combinations to compute the Shapley values. Another
model-agnostic approach is proposed by Ribeiro et al.~(2016)
\citep{ribeiro2016should}. LIME (Local Interpretable Model-agnostic
Explanations) explains individual predictions by approximating the model
locally with an interpretable model, often linear regression. While
LIME's flexibility and user-friendly nature contribute to its
popularity, it may exhibit less stability and consistency compared to
SHAP, primarily due to its reliance on local approximations. A third
notable model-agnostic method is permutation-based variable importance
\citep{fisher2019all}. It assesses how the model's prediction error
changes when the values of a feature are permuted. This technique offers
a global measure of feature importance applicable to any model. However,
Feature Permutation can be computationally intensive for
high-dimensional data and may not fully capture complex feature
interactions.

Other attribution methods have been developed specifically for certain
models, utilizing components inherent to their architectures, such as
gradients in neural networks. DeepLIFT, proposed by Shrikumar et
al.~(2017) \citep{shrikumar2017learning}, calculates importance scores
by comparing neuron activations to reference activations. Effective for
non-linear models, DeepLIFT captures complex feature interactions,
particularly beneficial for high-dimensional problems. It serves as a
faster alternative to SHAP values while maintaining interpretability.
Introduced by Bach et al.~(2015) \citep{bach2015pixel}, Layer-Wise
Relavance Propagation (LRP) decomposes the output of a neural network
recursively, layer by layer, assigning relevance to each input feature.
While originally developed for image classification, it has been adapted
for other tasks. LRP provides detailed attributions at the input feature
level, but can be computationally intensive for deep networks. IG,
developed by Sundararajan et al.~(2017)
\citep{sundararajan2017axiomatic}, attributes predictions of deep
networks by accumulating gradients along a straight-line path from a
baseline to the input. This method satisfies essential axioms such as
completeness and sensitivity, making it well-suited for neural network
models. IG leverages gradient information inherent in neural networks,
ensuring efficient computation and scalability to high-dimensional data.

While the majority of XAI investigations in real-world scenarios have
focused on classification problems, there are some publications
exploring the application of these techniques to regression tasks.
Recent studies have demonstrated the potential of XAI methods in several
domains where regression models are prevalent. Mamalakis et al.~(2023)
conducted a comprehensive analysis of feature attribution methods in
climate science \citep{Mamalakis2023}. Their work highlights the
importance of carefully selecting baselines when applying XAI techniques
to regression problems in geoscientific applications. The authors showed
that different baselines can lead to substantially different
attributions, both in magnitude and spatial patterns, emphasising the
need for careful consideration of baselines to avoid misidentifying
sources of predictability. In a different domain, Letzgus et al.~(2022)
examined two different use cases for feature attribution in regression
problems \citep{Letzgus2022}. The first involved age prediction from
facial images, demonstrating how XAI techniques can provide insight into
the features contributing to age estimates. The second focused on
molecular energy prediction in quantum chemistry, demonstrating the
versatility of XAI methods across different scientific disciplines.
Notably, these studies focus primarily on local feature mapping, aiming
to explain specific predictions rather than the overall behaviour of the
models. This approach allows for a granular understanding of how
individual features contribute to particular outcomes, which is
particularly valuable in critical domains where decision-making
processes need to be transparent and justifiable.

Even though feature attribution has not been extensively investigated in
the context of feature selection, initial work has begun to address this
topic. Zacharias et al.~(2022) employ SHAP for XGBoost to enhance the
interpretability of model preprocessing steps
\citep{zacharias2022designing}. Van Zyl et al.~(2023) utilize SHAP and
Grad-CAM for feature selection in Convolutional Neural Networks applied
to time series data \citep{van2024harnessing}. This work offers
promising results regarding improvements in model performance and
reduction in training time.

Table~\ref{tbl-feature-selection} provides an overview of the advantages
and disadvantages of different feature selection methods.

\begin{longtable}[]{@{}
  >{\raggedright\arraybackslash}p{(\columnwidth - 6\tabcolsep) * \real{0.2015}}
  >{\raggedright\arraybackslash}p{(\columnwidth - 6\tabcolsep) * \real{0.1791}}
  >{\raggedright\arraybackslash}p{(\columnwidth - 6\tabcolsep) * \real{0.3731}}
  >{\raggedright\arraybackslash}p{(\columnwidth - 6\tabcolsep) * \real{0.2463}}@{}}
\caption{Comparison of feature selection
methods.}\label{tbl-feature-selection}\tabularnewline
\toprule\noalign{}
\begin{minipage}[b]{\linewidth}\raggedright
Group
\end{minipage} & \begin{minipage}[b]{\linewidth}\raggedright
Example
\end{minipage} & \begin{minipage}[b]{\linewidth}\raggedright
Advantages
\end{minipage} & \begin{minipage}[b]{\linewidth}\raggedright
Disadvantages
\end{minipage} \\
\midrule\noalign{}
\endfirsthead
\toprule\noalign{}
\begin{minipage}[b]{\linewidth}\raggedright
Group
\end{minipage} & \begin{minipage}[b]{\linewidth}\raggedright
Example
\end{minipage} & \begin{minipage}[b]{\linewidth}\raggedright
Advantages
\end{minipage} & \begin{minipage}[b]{\linewidth}\raggedright
Disadvantages
\end{minipage} \\
\midrule\noalign{}
\endhead
\bottomrule\noalign{}
\endlastfoot
Filters & Chi-Square Test & Simple, fast, univariate. Suitable for
categorical data. & Ignores feature interactions. Not suitable for
continuous features directly (requires discretization). \\
Wrappers & Recursive Feature Elimination (RFE) & Considers interactions,
generally good performance & Computationally intensive \\
Embedding Methods & LASSO & Embedded in model training, can handle large
datasets & May not perform well if features are highly correlated \\
Feature attribution & SHAP & Stable, local accuracy, consistency & High
computational demand \\
Feature attribution & DeepLIFT & Fast, local explanations & Only
applicable to neural networks \\
Feature attribution & LRP (Layer-wise Relevance Propagation) & Provides
detailed insights into model decisions & Requires a specific type of
neural network \\
Feature attribution & Integrated Gradients & Theoretically sound, works
with any differentiable model & Requires multiple model evaluations,
high computational cost \\
\end{longtable}

\subsection{Neural Networks for Turbo
Machinery}\label{neural-networks-for-turbo-machinery}

Turbomachines are inherently complex systems characterized by highly
nonlinear physical processes. The vibration behavior of these machines
is influenced by numerous unknown variables, making accurate modeling
and analysis challenging. Additionally, the presence of measurement
noise further complicates the identification of underlying patterns and
behaviors. Therefore, the integration of Deep Learning and Machine
Learning into turbomachinery design has garnered significant attention
in recent years, with various studies exploring their potential to
enhance design efficiency and performance.

Michelassi and Ling (2021) \citep{michelassi2021challenges} discuss
recent trends in turbomachinery design methods that leverage both AI and
high-fidelity simulations to address the increasing demands for
efficiency, availability, and cost of ownership in energy conversion and
propulsion systems. One of the key components is the supplementation of
physical models with data-driven approaches. The authors emphasize that
AI approaches do not replace scientific expertise but rather build upon
it, combining data-driven methods with traditional scientific knowledge
to overcome limitations in available data within the turbomachinery
field.

Fei et al.~(2016) \citep{fei2016compressor} investigated the potential
of neural networks with Gaussian kernel functions to predict compressor
performance maps. This novel approach was compared with other
data-driven models. It was shown that the proposed network is beneficial
compared to the other approaches, especially for extrapolating
compressor performance maps from a small number of data samples.

Weber et al.~(2024) \citep{weber2024compressor} provide an in-depth
analysis of compressor blade vibrations in turbochargers, emphasizing
the importance of damping in mitigating high-cycle fatigue. They explore
the integration of AI techniques with traditional measurement methods to
advance turbocharger development. The study highlights the use of strain
gauges and blade tip timing for precise blade vibration measurement,
alongside a simplified method for predicting critical damping ratios
during the early design phase. In Focus, an AI-driven methodology
employing Deep Learning is used to assess genuine damping ratios from
experimental data. These approaches accelerate turbocharger development
by providing accurate data early in the design process, validating
numerical predictions, and enhancing confidence in these methodologies.
The AI-based damping estimator is particularly noteworthy, offering
engineers a practical tool for improving the reliability and efficiency
of turbocharger systems.

\section{Background of the Real World
Problem}\label{background-of-the-real-world-problem}

\subsection{High-Cycle Fatigue in Turbo Charging
Systems}\label{high-cycle-fatigue-in-turbo-charging-systems}

Modern turbocharging systems are critical for enhancing the efficiency
and power of engines. Beside the classic automotive domain this
technology is intensively developing in the marine and power industry.
There is a growing need to push the technological boundaries of these
systems, optimizing layout parameters beyond traditional applications.
The market demands turbochargers capable of achieving higher pressure
ratios and increased flow capacities while maintaining the highest
efficiencies. These features are crucial not only for the
competitiveness of the turbocharger itself but also for the engine
manufacturers who significantly benefit from this new generation of
turbochargers \citep{spengler2023high}.

However, these advancements come with substantial challenges. Components
such as compressors and turbines must withstand higher loads, requiring
robust designs that account for various fatigue conditions, including
low-cycle fatigue, thermo-mechanical fatigue, and especially high-cycle
fatigue (HCF)
\citep{weber2024compressor}\citep{spengler2023high}\citep{nicholas1999critical}.
As pressure levels rise, the vibrations get stronger, and the faster air
flow makes the conditions more complex, causing more stress on the
parts. To meet thermodynamic demands for efficiency and pressure, bladed
nozzle rings with minimal clearance to rotating parts are used. This
design choice focuses on the related excitation orders (EOs) and the
blade design itself to ensure an HCF-safe design. Additionally, the wide
variety of differing variation parts used in turbochargers to meet
specific medium- and low-speed engine requirements adds another layer of
complexity. This complexity is particularly pronounced in the area of
vibration analysis, where complex interactions must be thoroughly
understood. Engineers often focus on determining resonant frequencies
with their vibration behavior and assessing the effects of various
physical parameters
\citep{spengler2023high}\citep{castanier2006modeling}.

Traditionally, this process relies heavily on extensive numerical
simulations and physical testing, the latter of which can be
particularly costly and time-consuming. Due to the high cost of testing
these configurations, data-driven approaches are being explored to
complement and partially replace traditional methods in determining
HCF-safe designs. This involves predicting amplitudes, as discussed in
this paper.

\subsection{Description of Experiment Data}\label{sec-data}

\begin{figure}[H]

{\centering \includegraphics[width=0.7\textwidth,height=\textheight]{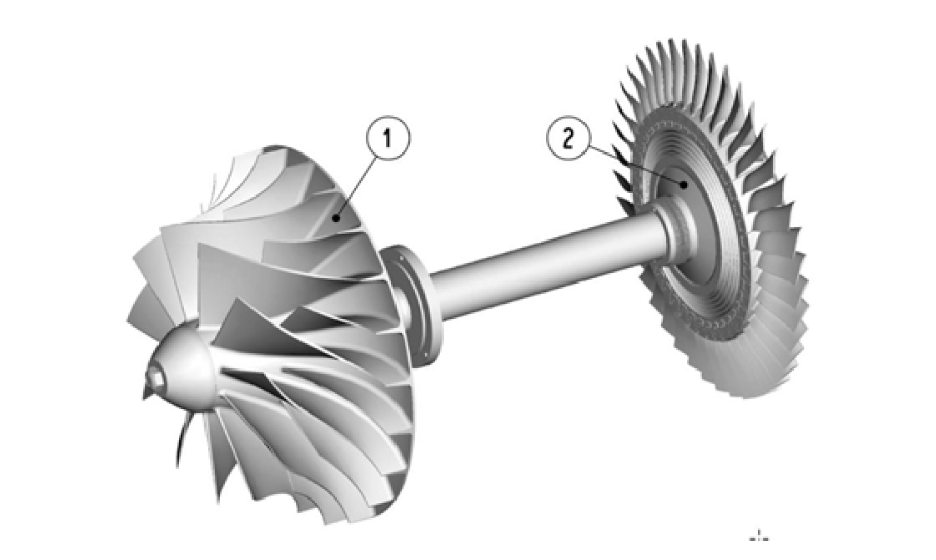}

}

\caption{Illustrative visualization of the turbine wheel (1) and
compressor wheel (2) in a turbocharging system. Critical vibrations
occur at the blades mounted on the wheels. (MAN Energy Solutions,
2023)\label{fig-blades}}

\end{figure}

The data was provided by MAN Energy Solutions. Due to confidentiality
and the high level of competition in this domain, we are not authorized
to publish the data. The data set is based on a test setup for
turbomachinery. The systems are subjected to different excitation
frequencies to identify critical oscillation behaviors. Stresses (or
amplitudes) at the blades are measured using strain gauges and analyzed
with appropriate software to calculate the stresses at the most critical
points. The calculated stress at these critical points is referred to as
the `upgraded amplitude'. This interim step is necessary to minimize the
number of strain gauges used. The positioning of the strain gauges is a
compromise to capture and upgrade many different eigenmodes of the
system. Otherwise, it would be necessary to attach a separate strain
gauge for each eigenmode, as the peaks never occur at the same location.
The aim of our experiments is to predict the amplitude levels based on
various process parameters measured during the test runs. An examplary
illustration of an turbine and compressor wheel of a turbo machine can
be seen in \ref{fig-blades}.

We divide the features that we use to predict the upgraded amplitude
into different subgroups. Most of the parameters are thermodynamic
variables, such as pressures and temperatures at various points in the
system. This group accounts for 67 features out of 86. Five features can
be summarized under the term `Quality Criteria', which contain various
quality measures for the numerical valorization of the amplitudes. We
summarize the information on modes, node diameters, and order under the
heading `Oscillations Characteristics'. In addition to the thermodynamic
process parameters, purely mechanical parameters are also measured.
These include compressor and motor speeds at the resonance points,
vibration frequencies of the turbine blades, and circumferential speeds
of the compressor wheel. The time stamp of the measurements represents a
separate category. The area ratio between the inlet of the compressor
wheel and the inlet of the bladed outlet diffusor, as well as the
throttle positions of the system, refer to the geometric setup. We have
defined the designations of the measurement series and strain gauges as
part of the experimental setup.

\section{Methods}\label{methods}

\subsection{Data Preprocessing}\label{data-preprocessing}

In our experiments, we utilize 86 features. The target variable we aim
to predict is the upgraded oscillation amplitude at the rotor blades of
the system. Most of the data consists of numerical values, except for
two features that contain categorical data: the labels of the strain
gauges and the measurement series. First, we remove data points that
contain NA entries or are labeled with insufficient measurement quality.
We also exclude data with implausible values, such as negative values
for modes or frequencies that are too low and were not part of the
measurement process. After this filtering process, we retain 27,857 data
samples.

After data cleaning, we encode the two categorical columns using
scikit-learn's implementation of an ordinal encoder, transforming them
into numerical values. The data is then scaled using our custom
implementation of a standard scaler. To avoid data leakage, we fit the
scaler by calculating the mean and standard deviation of our training
data, which comprises 70\% of the entire dataset. We then transform both
the training and test data using the fitted scaler. This approach
ensures that the test data remains unseen during the training process,
maintaining the integrity of our evaluation.

\subsection{Neural Network Architecture and Hyperparameter Tuning with
spotpython}\label{sec-tuning}

\subsubsection{Introduction to spotpython}\label{sec-spot}

Surrogate model-based optimization methods are widely used in the fields
of simulation and optimization. The Sequential Parameter Optimization
Toolbox (SPOT) was developed to address the need for robust statistical
analysis of simulation and optimization algorithms. It is freely
available on GitHub as a Python package, \texttt{spotpython}\footnote{https://github.com/sequential-parameter-optimization/}.
SPOT offers advanced methodologies for optimization (tuning), based on
classical regression techniques and analysis of variance. It includes
state-of-the-art tree-based models such as classification and regression
trees, and random forests, as well as Bayesian optimization models like
Gaussian processes, also known as Kriging. Additionally, SPOT allows for
the integration and combination of different meta-modeling approaches.

SPOT comes equipped with a sophisticated surrogate model-based
optimization method capable of handling both discrete and continuous
inputs. Furthermore, any model implemented in \texttt{scikit-learn} can
be seamlessly used as a surrogate within \texttt{spotpython}, providing
extensive flexibility to users.

SPOT also implements crucial techniques such as exploratory fitness
landscape analysis and sensitivity analysis. These features enable users
to comprehensively understand the performance and underlying behavior of
various algorithms, thus enriching algorithmic insights and enhancing
optimization outcomes. Hyperparameter tuning can be performed using
SPOT, which is particularly useful for optimizing neural network
architectures.

The \texttt{spot} loop consists of the following steps:

\begin{enumerate}
\def\labelenumi{\arabic{enumi}.}
\tightlist
\item
  Init: Build initial design \(X\)
\item
  Evaluate initial design on real objective \(f\): \(y = f(X)\)
\item
  Build surrogate: \(S = S(X,y)\)
\item
  Optimize on surrogate: \(X_0 =  \text{optimize}(S)\)
\item
  Evaluate on real objective: \(y_0 = f(X_0)\)
\item
  Impute (Infill) new points: \(X = X \cup X_0\), \(y = y \cup y_0\).
\item
  Goto 3.
\end{enumerate}

\subsubsection{The Neural-Network
Architecture}\label{the-neural-network-architecture}

A neural network class for regression tasks implemented in
\texttt{spotpython} is used for the neural network for both dummy data
and real-world experiments. The network is a multi-layer perceptron
(MLP) that is constructed using the Python frameworks PyTorch and
PyTorch-Lightning
\citep{NEURIPS2019_9015}\citep{Falcon_PyTorch_Lightning_2019}. This
network consists of four hidden layers. The input layer has the same
dimension as the feature space. The optimal number of neurons in the
first hidden layer is determined through hyperparameter tuning. The
second and third hidden layers have half of the neurons of the first
hidden layer, while the third hidden layer has only a quarter of the
neurons of the first hidden layer. After each of these layers, there is
a dropout layer. The output layer of the network is a linear layer with
one neuron, as we want to predict one target variable. During
hyperparameter optimization, the activation functions of the hidden
layers, the batch size, the dropout probabilities of the dropout layers,
the number of training epochs, and the number of neurons in the first
hidden layer are optimized. In addition, the best optimizer and the
number of epochs that the training is continued without improvement in
performance are determined. Instead of tuning the learning rate itself,
a multiplication factor for the learning rate is tuned in
\texttt{spotpython}. This is useful because the learning rates for
different optimizers differ. For example, the learning rate for the
Adagrad optimizer results from the \(learning rate \times 0.01\), while
for the Adam optimizer it results from the formula
\(learning rate \times 0.001\). Further details can be found in the
\texttt{spotpython} source code. For all tuning runs related to this
article, surrogate-based optimization is used. The surrogate model is a
Gaussian process model built on 10 initial evaluations. During the
tuning process, 30 additional evaluations are performed. The boundaries
of the numerical hyperparameters are shown in Table~\ref{tbl-tuning},
and those for the categorical hyperparameters are shown in
Table~\ref{tbl-tuning2}.

\begin{longtable}[]{@{}
  >{\raggedright\arraybackslash}p{(\columnwidth - 10\tabcolsep) * \real{0.2051}}
  >{\raggedright\arraybackslash}p{(\columnwidth - 10\tabcolsep) * \real{0.1282}}
  >{\raggedright\arraybackslash}p{(\columnwidth - 10\tabcolsep) * \real{0.1282}}
  >{\raggedright\arraybackslash}p{(\columnwidth - 10\tabcolsep) * \real{0.1282}}
  >{\raggedright\arraybackslash}p{(\columnwidth - 10\tabcolsep) * \real{0.1282}}
  >{\raggedright\arraybackslash}p{(\columnwidth - 10\tabcolsep) * \real{0.2821}}@{}}
\caption{Numerical hyperparameters and their boundaries. The \(2^x\)
transformation indicates that the integers describing the parameters are
used as exponents of 2.}\label{tbl-tuning}\tabularnewline
\toprule\noalign{}
\begin{minipage}[b]{\linewidth}\raggedright
Parameter
\end{minipage} & \begin{minipage}[b]{\linewidth}\raggedright
Lower Bound
\end{minipage} & \begin{minipage}[b]{\linewidth}\raggedright
Upper Bound
\end{minipage} & \begin{minipage}[b]{\linewidth}\raggedright
Type
\end{minipage} & \begin{minipage}[b]{\linewidth}\raggedright
Transformation
\end{minipage} & \begin{minipage}[b]{\linewidth}\raggedright
Description
\end{minipage} \\
\midrule\noalign{}
\endfirsthead
\toprule\noalign{}
\begin{minipage}[b]{\linewidth}\raggedright
Parameter
\end{minipage} & \begin{minipage}[b]{\linewidth}\raggedright
Lower Bound
\end{minipage} & \begin{minipage}[b]{\linewidth}\raggedright
Upper Bound
\end{minipage} & \begin{minipage}[b]{\linewidth}\raggedright
Type
\end{minipage} & \begin{minipage}[b]{\linewidth}\raggedright
Transformation
\end{minipage} & \begin{minipage}[b]{\linewidth}\raggedright
Description
\end{minipage} \\
\midrule\noalign{}
\endhead
\bottomrule\noalign{}
\endlastfoot
l1 & 5 & 9 & Integer & \(2^x\) & Number of neurons in the first hidden
layer \\
epochs & 5 & 10 & Integer & \(2^x\) & Number of training epochs \\
batch\_size & 3 & 6 & Integer & \(2^x\) & Batch size \\
dropout\_prob & 0.005 & 0.25 & Float & - & Dropout probability for the
dropout layer \\
lr\_mult & 0.25 & 5.0 & Float & - & Multiplier for the learning rate of
the optimizer \\
patience & 3 & 5 & Integer & \(2^x\) & Number of epochs the training
continues without improvement \\
\end{longtable}

\begin{longtable}[]{@{}
  >{\raggedright\arraybackslash}p{(\columnwidth - 4\tabcolsep) * \real{0.2308}}
  >{\raggedright\arraybackslash}p{(\columnwidth - 4\tabcolsep) * \real{0.3846}}
  >{\raggedright\arraybackslash}p{(\columnwidth - 4\tabcolsep) * \real{0.3846}}@{}}
\caption{Categorical hyperparameters and their
levels.}\label{tbl-tuning2}\tabularnewline
\toprule\noalign{}
\begin{minipage}[b]{\linewidth}\raggedright
Parameter
\end{minipage} & \begin{minipage}[b]{\linewidth}\raggedright
Levels
\end{minipage} & \begin{minipage}[b]{\linewidth}\raggedright
Description
\end{minipage} \\
\midrule\noalign{}
\endfirsthead
\toprule\noalign{}
\begin{minipage}[b]{\linewidth}\raggedright
Parameter
\end{minipage} & \begin{minipage}[b]{\linewidth}\raggedright
Levels
\end{minipage} & \begin{minipage}[b]{\linewidth}\raggedright
Description
\end{minipage} \\
\midrule\noalign{}
\endhead
\bottomrule\noalign{}
\endlastfoot
optimizer & Adadelta, Adamax, Adagrad & The optimizer that is during for
backpropagation \\
act\_fn & ReLU, LeakyReLU & The activation function for the layers of
the neural network \\
\end{longtable}

\subsection{Integrated Gradients}\label{integrated-gradients}

Introduced by Sundararajan et al.~(2017), IG is a feature attribution
method that aims to attribute the prediction of a deep learning model to
its input features by integrating gradients along the path from a
baseline input to the actual input \citep{sundararajan2017axiomatic}.
The key idea is to compute the integral of the gradients of the model's
output with respect to the input features, taken along a straight path
from a baseline (typically an input with no information, such as a black
image or a zero vector) to the actual input. The algorithm fulfills the
two axioms of sensitivity and implementation invariance. Sensitivity
means that if an input feature changes the model's prediction, the
attribution for that feature should be non-zero. Implementation
invariance means that if two models produce the same prediction for the
same input, the attribution values should also be the same. IG can be
described by the following formula:

\[
IG_{i}(x) = (x_{i}-x_{i}') \times \int_{\alpha=0}^{1} \frac{\partial F(x' + \alpha \times (x - x'))}{\partial x_i} \, d\alpha
\]

where:

\begin{itemize}
\item
  \(x\) is the input for which the attributions are computed.
\item
  \(x'\) is a baseline input.
\item
  \(\alpha\) is a scalar that scales the interpolation between the
  baseline input \(x'\) and the input \(x\).
\item
  \(\frac{\partial F(x)}{\partial x_i}\) is the gradient of the model
  output \(F(x)\) with respect to the \(i\)-th input feature.
\end{itemize}

Since calculating the integral directly can be computationally
intensive, it is often approximated using methods like the Riemann sum
or Gauss-Legendre quadrature rule. For our specific problem of selecting
the most relevant features to improve neural network regression
performance, we argue that IG is the most suitable approach. This choice
is supported by several key advantages that IG offers in the context of
complex neural network models. Firstly, IG stands on a solid theoretical
foundation. It satisfies important axioms such as completeness and
sensitivity, ensuring that the attributions it provides are both
reliable and consistent \citep{sundararajan2017axiomatic}
\citep{wang2024gradient}. This theoretical soundness is crucial when
dealing with the intricate decision-making processes of neural networks.
Secondly, IG demonstrates excellent scalability, a critical factor in
our high-dimensional regression tasks. Its computational efficiency
allows it to handle large-scale datasets without compromising
performance, making it particularly well-suited for complex regression
problems that often involve numerous input features. As a model-specific
method, IG provides insights that are directly tied to the neural
network's internal workings. This characteristic allows for more
accurate and relevant attributions compared to model-agnostic
approaches, which may not capture the nuances of the neural network's
decision-making process. Furthermore, IG's gradient-based approach
aligns well with the architecture of neural networks. By leveraging the
gradient information that is readily available in these models, IG can
efficiently perform attributions by quantifying how infinitesimal
changes in input features affect the prediction, making it a natural fit
for our neural network-based regression tasks \citep{wang2024gradient}.
Lastly, IG excels in handling non-linearity, a common characteristic of
complex neural network models. Unlike simpler attribution methods, IG
can effectively capture and represent non-linear relationships between
inputs and outputs, providing a more comprehensive understanding of
feature importance in our regression tasks.

\subsection{KernelSHAP}\label{kernelshap}

Kernel SHAP is an estimation method for Shapley values using a weighted
linear regression. The Shapley value for a feature (i) can be
approximated as \citep{lundberg2017unified}:

\[
\phi_i = \sum_{z' \subseteq x'} \frac{|\mathcal{Z}| - 1}{\binom{|\mathcal{Z}|}{|z'|} |z'| (|\mathcal{Z}| - |z'|)} \cdot \left[ f(z' \cup x_i) - f(z') \right]
\]

where:

\begin{itemize}
\tightlist
\item
  \(\phi_i\) is the Shapley value for feature \(i\).
\item
  \(z'\) represents a subset of the feature set \(x'\).
\item
  \(|\mathcal{Z}|\) is the total number of features.
\item
  \(|z'|\) is the number of features in subset \(z'\).
\item
  \(f(z' \cup x_i)\) is the model output when feature \(i\) is added to
  subset \(z'\).
\item
  \(f(z')\) is the model output for subset \(z'\).
\end{itemize}

We use this approach to compare the feature selection abilities with our
approach that relies on IG. We chose this method because Shapley values
are one of the most well-known XAI methods. For our real-world
experiments, where we have an initial data set of 86 features, computing
the classical Shapley values would be computationally unfeasible, due to
their exponential time complexity \citep{strumbelj2010efficient}.
Therefore, we decided to use KernelShap as an approximator.

\section{Experimental Setup}\label{experimental-setup}

To underline the reliability of our methods, we start by applying IG to
a dummy dataset that we generated. This allows us to determine the
ground truth regarding the importance of the variables and to test
whether our approach can effectively filter out unimportant features.
Afterwards, we continue by using IG in our feature selection pipeline
for our real-world experiment.

\subsection{Dummy}\label{dummy}

To generate the dummy data, we have written a function based on a linear
model. The coefficients are generated using a uniform distribution. To
resemble the real-world experiment, 86 coefficients are generated. It is
assumed that 29 of the coefficients are positively correlated with the
target variable (coefficients between 0.1 and 1), and another 29 are
negatively correlated (coefficients between -0.1 and -1). The remaining
28 coefficients have no significance for predicting the target variable
and therefore have a value of 0. The coefficients are multiplied by
randomly generated X values, which add up to the target variable.
Analogous to our real-world experiments, 27,857 data points are
generated in this manner. No feature interactions are considered, as
this information is not available for the real-world data either. The
source code for generating the data, as well as the data itself, can be
found on GitHub\footnote{https://github.com/ahinterl94-th/data\_generator}.
In the dummy experiments, the aim is to filter out these insignificant
values using IG. Therefore we start by optimizing the neural network
based on the tuning setup as described in Section~\ref{sec-tuning},
followed by the IG analysis.

\subsection{Real World Experiment}\label{real-world-experiment}

\subsubsection{Baseline with linear
Model}\label{baseline-with-linear-model}

For some engineering problems, classical regression models are better
suited than more advanced algorithms. Classical models also have the
advantage that the results are often much easier to interpret. For
complex problems involving non-linearities and noisy data, more complex
black box models such as neural networks are the better choice
\citep{alzubaidi2021review}\citep{lecun2015deep}. To ensure that our
problem is non-trivial and necessitates advanced methods, such as a
neural network, we first attempt to predict the oscillation amplitudes
using simple linear regression, employing the \texttt{sklearn}
implementation \citep{scikit-learn}. We utilize all 86 features, as
described in Chapter X, and split the data into 70\% training and 30\%
test sets. Additionally, we perform five fold cross-validation to ensure
robust and reliable results.

\subsubsection{Feature Selection Pipeline}\label{sec-fsp}

\begin{figure}[H]

{\centering \includegraphics[width=0.7\textwidth,height=\textheight]{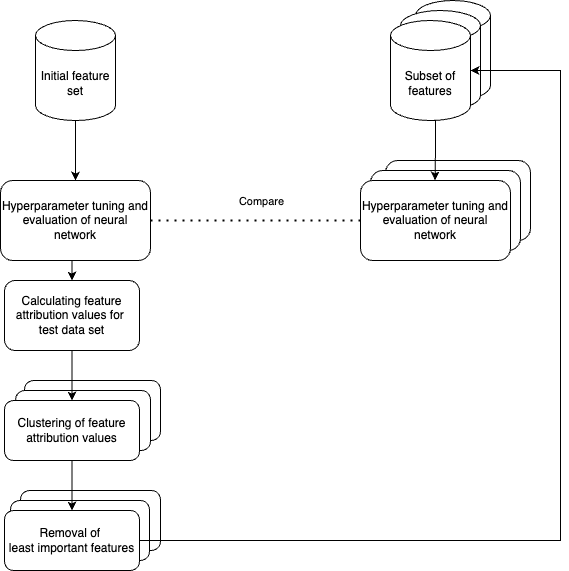}

}

\caption{Process for feature selection based on feature attribution and
clustering. The process starts with tuning a neural network using all
available features. The tuned network is then analyzed using integrated
gradients to calculate attribution values. Next, these attribution
values are clustered with varying k values. By removing the least
important cluster from each clustering, different new feature sets are
created. A neural network is subsequently tuned for each of these
feature sets. The results are compared with each other and with the
network based on the initial feature set. \label{fig-process}}

\end{figure}

Our novel feature selection process can be described as a data-driven
reverse engineering approach. First, the neural network is tuned, taking
into account all available features. This ensures that we have a
suitable architecture for the corresponding data. This step is
particularly important because feature attribution methods, such as IG,
examine the extent to which features contribute to the prediction of the
target variables. Therefore, the trustworthiness of these methods
depends on the performance of the network. After optimizing the
hyperparameters, the model is tested using a k-fold cross-validation
with five folds. Model performance for all experiments in this article
is measured using Mean Squared Error (MSE). This metric is particularly
suitable for our use case, where predicting oscillation amplitudes
accurately, including extreme values or outliers, is crucial. MSE is the
preferred choice because it penalizes outliers more heavily during
training, thus encouraging the model to improve its predictions for
these extreme values. The attribution values are calculated using an IG
implementation that uses a Gaussian Legendre quadrature with 50
approximation steps from the Python package Captum
\citep{kokhlikyan2021captum}. The algorithm computes local attribution
values for each sample in the dataset. To make a global statement about
the importance of the features, the mean of all local attribution values
is determined. The baseline can be viewed as a state where all features
are absent, which in our case means that the system is in a state where
it does not vibrate at all. Since we aim to determine the features
generally relevant for predicting upgraded vibration amplitudes, we set
the baseline for IG to a null vector. Changing the baseline to a state
of expected vibration would be more appropriate if we wanted to explain
individual predictions that deviate from the expected behavior. After
aggregating the attribution values, they are grouped using a K-Means
clustering algorithm \citep{lloyd1982least}. The number of clusters
(\(k\)) is varied to categorize the features into different importance
classes. Following each clustering procedure, the cluster containing the
least important features is removed from the dataset. This process
results in \(n\) distinct sub-datasets, where \(n\) is the number of
K-Means evaluations with different \(k\) values. This approach
represents a data-driven method for filtering features based on their
importance in predicting the target variable. Consequently, there are no
fixed limits on the number of features to include or exclude. Following
the feature selection process, the models are re-tuned using each of the
distinct feature subsets. This crucial step ensures a fair comparison
among models based on different feature combinations. After the tuning,
each model is evaluated using the k-fold cross-validation with five
folds. By doing so, we can effectively validate the efficancy of our
feature selection methodology. The whole process of the feature
selection process using IG and K-Means is visualized in Figure
\ref{fig-process}.

\subsubsection{Feature Selection Validation
Experiments}\label{feature-selection-validation-experiments}

As this work serves as a proof of concept for utilizing feature
attribution for explainable feature selection, we aim to validate our
approach by conducting four additional comparative experiments.

In the first experiment, we perform a Cross-Check by taking the best
performing feature subset identified previously and then tune and
evaluate a network using the remaining features from the initial feature
set, which were identified as having the lowest information density. It
is hypothesized that the performance of this model should be
significantly worse compared to the model utilizing the important
features. This procedure is designed to verify that the feature
importance methods have a meaningful impact and do not exhibit random
behavior.

The second and third experiments employ two classical feature selection
methods. In one approach, we use the Pearson correlation to select the
top \(n\) features with the highest correlation to the target variable,
where \(n\) corresponds to the number of features in the best-performing
subset from the experiment described in Section~\ref{sec-fsp}. The other
classical method is an embedded approach: Lasso regression. This linear
model is a regularization technique that performs both feature selection
and coefficient shrinkage by adding an L1 penalty term to the linear
regression cost function. In this study, we opted for Lasso regression
over Ridge regression for feature selection due to its ability to shrink
coefficients to exactly zero, thereby effectively eliminating less
important features, which aligns with our goal of identifying the most
influential predictors in our high-dimensional dataset
\citep{tibshirani1996regression}\citep{filzmoser2012review}.

The final experiment involves using KernelShap to identify the \(n\)
most important features. This serves as a comparison of the feature
selection capabilities of IG with another attribution method, allowing
us to evaluate the robustness and effectiveness of our feature selection
approach.

\section{Results and Discussion}\label{results-and-discussion}

The ``Results and Discussion'' chapter is divided into two sections: the
results related to the dummy data and the results of the real-world
experiments. The first section validates the ability of IG to detect the
most important features based on the known ground truth of the dummy
data. The second section discusses the outcomes of the real-world
experiments.

\subsection{Results and Discussion of the Dummy Data
Experiments}\label{results-and-discussion-of-the-dummy-data-experiments}

\begin{figure}[H]

{\centering \includegraphics{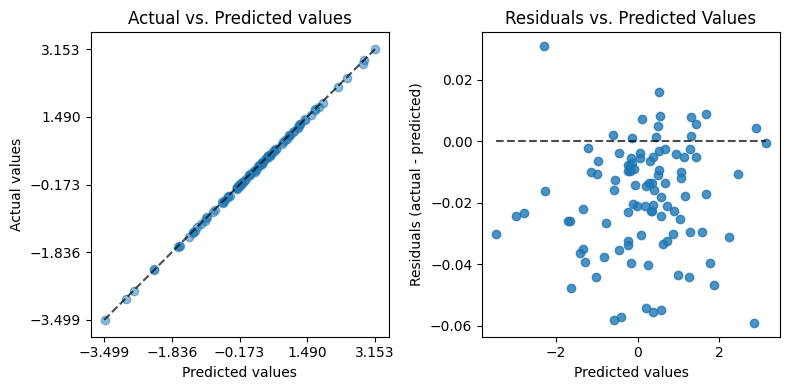}

}

\caption{Performance of the tuned neural network on the dummy data. The
left side shows the comparison between actual and predicted values. The
right graph shows the residuals between of the predicted values. The
graphs indicate that the network accurately captures the behavior of the
dummy data. \label{dummy_nn}}

\end{figure}

Figure \ref{dummy_nn} demonstrates the neural network's ability to
capture the behavior of the dummy data. This is visualized by comparing
the actual and predicted values, as well as by showing the residuals of
the predicted values. Due to the simple linear characteristics of the
dummy data, the network predicts the target values with high accuracy,
as indicated by the very small residuals compared to the target values.
The average MSE after cross-validation is 0.00073. Ensuring suitable
performance is crucial because the quality of the attribution methods
highly depends on the network's prediction accuracy.

\begin{figure}[H]

{\centering \includegraphics{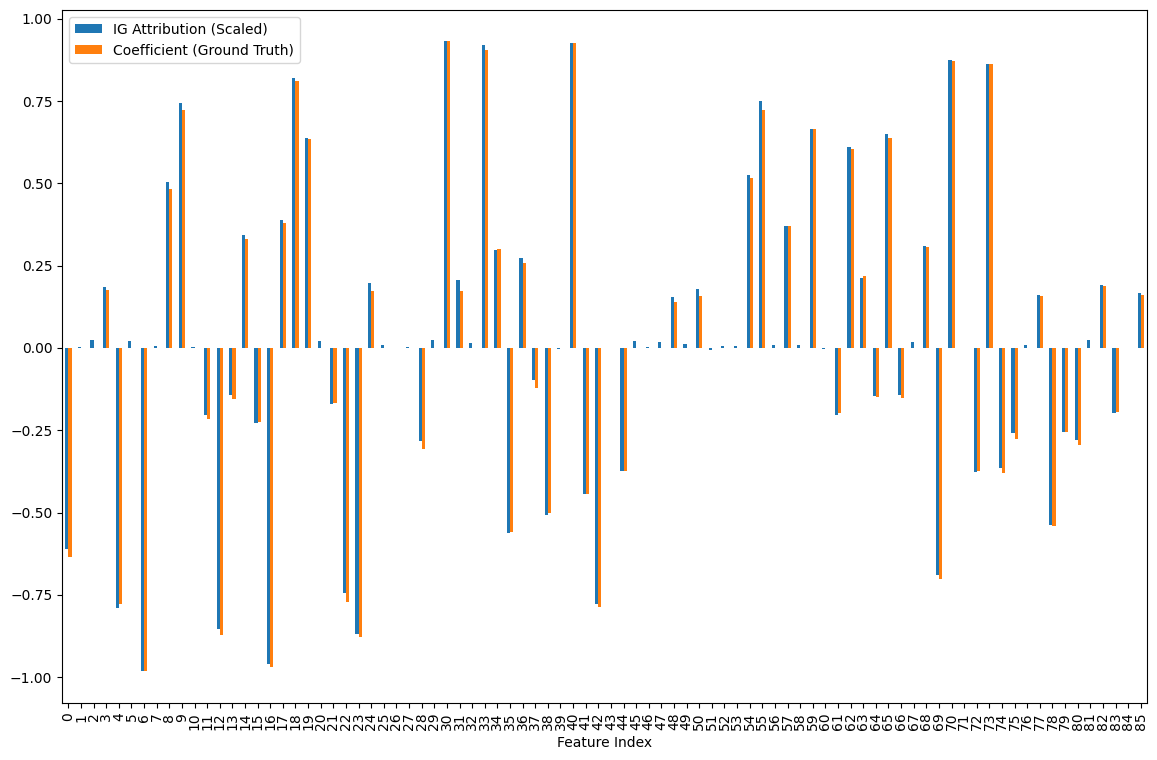}

}

\caption{Comparison between scaled IG feature attribution values and the
coefficients of the linear model used to generate the dummy data. The
attribution values are scaled between the minimum and maximum values of
the coefficients to ensure the same ranges for the comparison. It can be
seen that coefficients and the attribution values follow a similiar
pattern. \label{g_truth}}

\end{figure}

After evaluating the neural network, the IG analysis is performed.
Figure \ref{g_truth} shows the comparison between the scaled attribution
values (blue bars) and the coefficients of the linear model (orange
bars) used to generate the dummy data. The attribution values are scaled
between the minimum and maximum values of the coefficients using a
min-max scaler from \texttt{sklearn} \citep{scikit-learn}. This ensures
the comparability of the two sets of values. It can be seen that the two
bar plots follow a similar pattern for each feature. IG successfully
detected all relevant coefficients by assigning significantly higher
values to these features compared to those with a coefficient of zero.
Even the order of importance is correct. Features based on coefficients
of zero receiving negligible attribution values. These findings indicate
that IG is a suitable approach for assessing the importance of features
in high-dimensional datasets for regression tasks. However, it should be
noted that the dummy data does not include feature interactions or
non-linear behavior.

\subsection{Results and Discussion Real World
Experiment}\label{results-and-discussion-real-world-experiment}

The linear model, which serves as a baseline for our real-world
experiments, achieves an average MSE of 32,993 across five folds during
cross-validation. This performance is significantly worse than that of
our initial neural network, which achieves an average MSE of 8.718. This
result demonstrates that employing neural networks to capture the
complex non-linear patterns inherent in the underlying real-world
problem is a more effective approach.

\begin{figure}[H]

{\centering \includegraphics{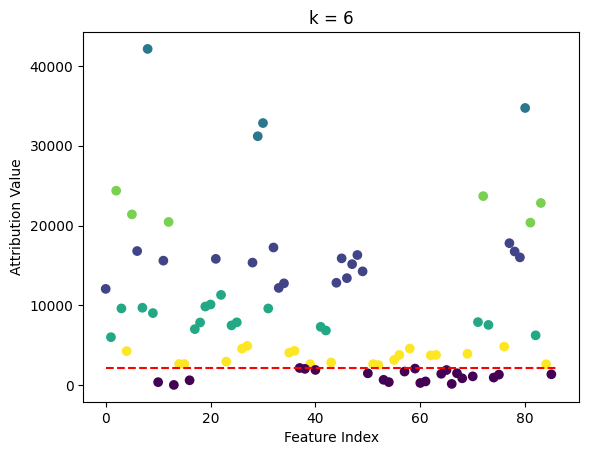}

}

\caption{K-Means clustering for the Integrated Gradient based feature
attribution values. The attribution values were calculated for a tuned
neural network across all 86 input features. This visualization
exemplifies the clustering process for K-Means with \(k = 6\). The
points on and below the red dotted line correspond to the cluster with
the lowest attribution values, which are removed during the feature
selection process. \label{clustering2}}

\end{figure}

When applying our feature selection pipeline, the attribution values,
which are based on the IG analysis of the initially tuned network with
all features, were clustered several times. The value of \(k\) was
changed for each clustering. In total, clustering was performed nine
times with \(k\) values ranging from two to ten. For each clustering
run, the cluster with the lowest attribution values, and therefore the
lowest relevance for the prediction, was removed from the initial
feature set. An examplary plot of the clustering process for \(k = 6\)
can be seen in Figure \ref{clustering2}. The points below and on the red
dotted line correspond to the lowest attribution values. The features
that are related to these points are removed during the feature
selection process. The visualization of all clustering runs for all
different \(k\)-values can be seen in the appendix. This visualizations
show that these clusters are identical for \(k\) values four and five,
as well as six to ten. This reduces the number of tunings and
evaluations performed with different feature sets from ten to five
(including the initial evaluation with all features).

Table Table~\ref{tbl-res1} shows the results of the evaluations of the
experiments in relation to the different \(k\) values and the associated
number of features. It can be seen that the initial tuning achieves an
average MSE of 8.718 across all cross-validations. However, the standard
deviation of the MSE values for the individual cross-validations is
higher here than in the evaluations of the experiments with reduced
feature sets. This suggests a lack of robustness in the predictions. For
the evaluations in which the number of features is reduced to less than
half of the initial feature set, the average MSE increases. This
indicates that with this drastic reduction of the feature set, relevant
information for predicting the vibration amplitudes was also removed
from the input data. The results of the predictions with 44 features
show, on average, a minimally lower MSE than the results of the
prediction with all 86 input variables. However, the standard deviation
is significantly lower for the 44 features. This indicates that
disturbance variables were filtered out, allowing the model to make more
stable predictions. The best performance in the experiments is achieved
with 64 features. A significantly better MSE is achieved here than with
the full data set. Additionally, the standard deviation is minimized in
this evaluation. This shows that IG could detect the features with the
highest information density and that the clustering process effectively
eliminated less relevant information and disturbances.

\begin{longtable}[]{@{}
  >{\raggedright\arraybackslash}p{(\columnwidth - 18\tabcolsep) * \real{0.1609}}
  >{\raggedright\arraybackslash}p{(\columnwidth - 18\tabcolsep) * \real{0.1034}}
  >{\raggedright\arraybackslash}p{(\columnwidth - 18\tabcolsep) * \real{0.0920}}
  >{\raggedright\arraybackslash}p{(\columnwidth - 18\tabcolsep) * \real{0.0920}}
  >{\raggedright\arraybackslash}p{(\columnwidth - 18\tabcolsep) * \real{0.0920}}
  >{\raggedright\arraybackslash}p{(\columnwidth - 18\tabcolsep) * \real{0.0920}}
  >{\raggedright\arraybackslash}p{(\columnwidth - 18\tabcolsep) * \real{0.0920}}
  >{\raggedright\arraybackslash}p{(\columnwidth - 18\tabcolsep) * \real{0.0920}}
  >{\raggedright\arraybackslash}p{(\columnwidth - 18\tabcolsep) * \real{0.0920}}
  >{\raggedright\arraybackslash}p{(\columnwidth - 18\tabcolsep) * \real{0.0920}}@{}}
\caption{The results of the cross-validation depend on the different
subsets of features that are selected using IG and K-Means. The table
presents the performance values as MSE, along with the selected number
of clusters and the resulting number of features. Additionally, it shows
the standard deviation of the MSE across all cross validations. In some
clustering experiments, the same number of resulting features is
obtained, as the cluster containing the least important features remains
unaffected by variations in the \(k\)
values.}\label{tbl-res1}\tabularnewline
\toprule\noalign{}
\begin{minipage}[b]{\linewidth}\raggedright
\(n\) Features
\end{minipage} & \begin{minipage}[b]{\linewidth}\raggedright
\(k\) Cluster
\end{minipage} & \begin{minipage}[b]{\linewidth}\raggedright
Fold 1 MSE
\end{minipage} & \begin{minipage}[b]{\linewidth}\raggedright
Fold 2 MSE
\end{minipage} & \begin{minipage}[b]{\linewidth}\raggedright
Fold 3 MSE
\end{minipage} & \begin{minipage}[b]{\linewidth}\raggedright
Fold 4 MSE
\end{minipage} & \begin{minipage}[b]{\linewidth}\raggedright
Fold 5 MSE
\end{minipage} & \begin{minipage}[b]{\linewidth}\raggedright
Mean MSE
\end{minipage} & \begin{minipage}[b]{\linewidth}\raggedright
SD MSE
\end{minipage} & \begin{minipage}[b]{\linewidth}\raggedright
\end{minipage} \\
\midrule\noalign{}
\endfirsthead
\toprule\noalign{}
\begin{minipage}[b]{\linewidth}\raggedright
\(n\) Features
\end{minipage} & \begin{minipage}[b]{\linewidth}\raggedright
\(k\) Cluster
\end{minipage} & \begin{minipage}[b]{\linewidth}\raggedright
Fold 1 MSE
\end{minipage} & \begin{minipage}[b]{\linewidth}\raggedright
Fold 2 MSE
\end{minipage} & \begin{minipage}[b]{\linewidth}\raggedright
Fold 3 MSE
\end{minipage} & \begin{minipage}[b]{\linewidth}\raggedright
Fold 4 MSE
\end{minipage} & \begin{minipage}[b]{\linewidth}\raggedright
Fold 5 MSE
\end{minipage} & \begin{minipage}[b]{\linewidth}\raggedright
Mean MSE
\end{minipage} & \begin{minipage}[b]{\linewidth}\raggedright
SD MSE
\end{minipage} & \begin{minipage}[b]{\linewidth}\raggedright
\end{minipage} \\
\midrule\noalign{}
\endhead
\bottomrule\noalign{}
\endlastfoot
86 & - & 7,705 & 8,250 & 11,428 & 9,402 & 6,807 & \textbf{8,718} & 1.594
& \\
27 & 2 & 11,94 & 9,601 & 8,716 & 10,156 & 10,387 & \textbf{10,160} &
1.060 & \\
33 & 3 & 8,30 & 10,282 & 11,379 & 9,729 & 8,155 & \textbf{9,570} & 1.218
& \\
44 & 4, 5 & 9,336 & 8,474 & 7,670 & 8,667 & 8,209 & \textbf{8,471} &
0.547 & \\
64 & 6-10 & 6,605 & 5,477 & 5,524 & 5,627 & 5,576 & \textbf{5,762} &
0.425 & \\
\end{longtable}

\begin{figure}[H]

{\centering \includegraphics{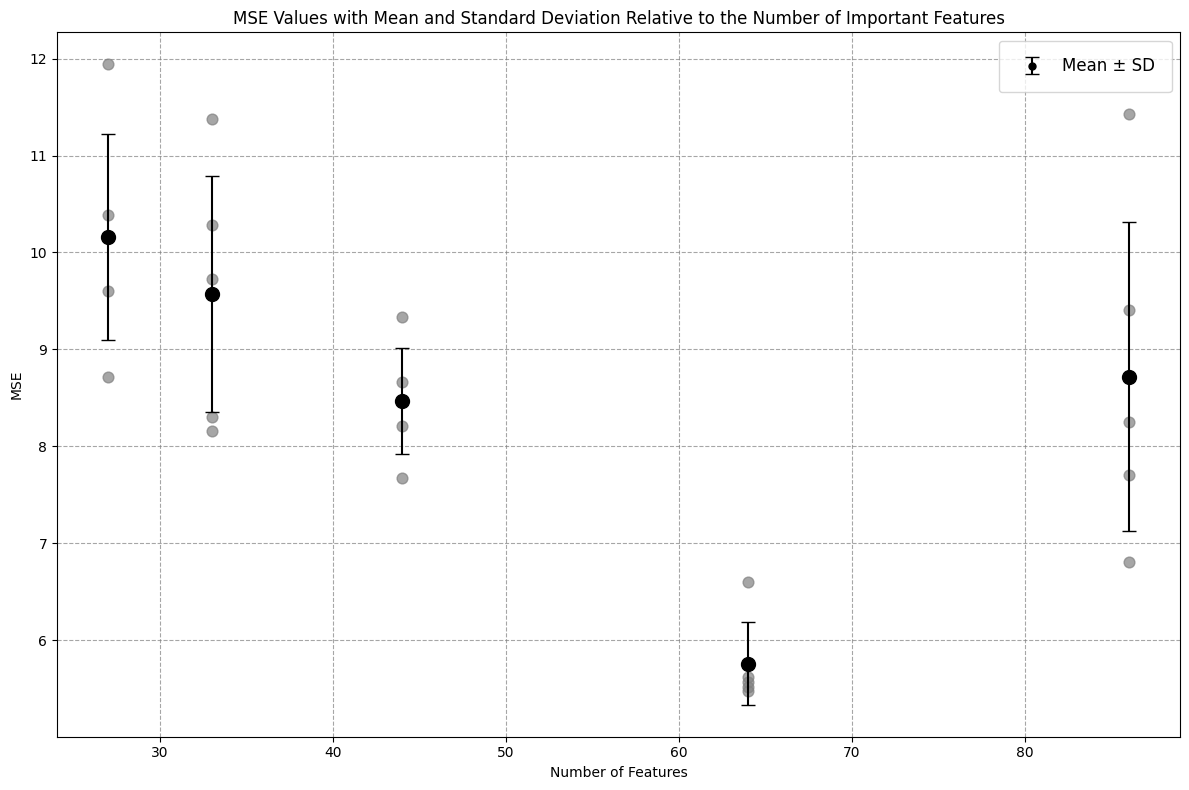}

}

\caption{Visualization of MSE values for predicting upgraded amplitudes
based on the number of important features determined by IG and used in
the neural network. The light grey points represent the actual MSE
values across all cross-validations for each feature set. The black dots
indicate the mean MSE, while the black lines represent the standard
deviation of the MSE values for each feature set. \label{fig-mse}}

\end{figure}

This observation is further supported by Figure \ref{fig-mse}. It
displays the individual MSE values (grey dots), along with the mean
(black dots) and standard deviations (black lines) of the MSE values,
based on the number of \(n\) most important features used. It is evident
that both the mean MSE and the variability of the MSE values are
minimized when using the 64 most important features.

\subsubsection{Results and Discussion of the Feature Selection
Validation
Experiments}\label{results-and-discussion-of-the-feature-selection-validation-experiments}

The results of the cross-validation experiments suggest that our
approach outperforms the other methods used for comparison in the
feature selection process. The experiment using the least important
features identified by our approach yields a MSE nearly three times
higher than that of the 64 most significant features determined by IG
(see Table~\ref{tbl-res2}). This outcome serves as a plausibility check
for our method, confirming its effectiveness. Utilizing Pearson
correlation for selecting the top 64 informative features results in
poor performance. This can be explained by the fact that all Pearson
correlation values are below 0.15. It is important to note that Pearson
correlation only accounts for linear relationships, which appears
inadequate for accurately capturing the complexities of a real-world
dataset. Lasso Regression, as a standard feature selection technique,
performs better than Pearson correlation but does not achieve as low an
MSE as the IG approach. However, the standard deviation of the results
is relatively low, indicating that the algorithm successfully captures
global importances. By introducing the regularization term, Lasso
Regression builds a more stable model compared to other approaches. The
KernelShap method serves as a direct competitor to IG, as both are
feature attribution techniques. The results indicate that the neural
network utilizing the 64 most important features identified by
KernelShap performs better than those relying on classical feature
selection approaches. However, the average MSE is higher than that of
the IG experiment, and the standard deviation is more than twice as
high. The results underline our assumption that the feature selection
loop based on IG is a suitable approach for identifying the most
important features. IG considers the gradient information of the
network, which is essential for the algorithm's decision-making process.
This can be particularly beneficial when working with high-dimensional
and complex data in the domain of neural networks, compared to other
feature selection methods.

\begin{longtable}[]{@{}
  >{\raggedright\arraybackslash}p{(\columnwidth - 16\tabcolsep) * \real{0.2593}}
  >{\raggedright\arraybackslash}p{(\columnwidth - 16\tabcolsep) * \real{0.0988}}
  >{\raggedright\arraybackslash}p{(\columnwidth - 16\tabcolsep) * \real{0.0988}}
  >{\raggedright\arraybackslash}p{(\columnwidth - 16\tabcolsep) * \real{0.0988}}
  >{\raggedright\arraybackslash}p{(\columnwidth - 16\tabcolsep) * \real{0.0988}}
  >{\raggedright\arraybackslash}p{(\columnwidth - 16\tabcolsep) * \real{0.0988}}
  >{\raggedright\arraybackslash}p{(\columnwidth - 16\tabcolsep) * \real{0.0988}}
  >{\raggedright\arraybackslash}p{(\columnwidth - 16\tabcolsep) * \real{0.0741}}
  >{\raggedright\arraybackslash}p{(\columnwidth - 16\tabcolsep) * \real{0.0741}}@{}}
\caption{Results of the comparison between the best performing feature
set, determined by IG and K-Means, and the other cross-check
experiments. It is shown that feature selection techniques based on
feature attribution seem to be beneficial in comparison to the classical
feature selection approaches.}\label{tbl-res2}\tabularnewline
\toprule\noalign{}
\begin{minipage}[b]{\linewidth}\raggedright
Experiment
\end{minipage} & \begin{minipage}[b]{\linewidth}\raggedright
\(n\) Features
\end{minipage} & \begin{minipage}[b]{\linewidth}\raggedright
Fold 1 MSE
\end{minipage} & \begin{minipage}[b]{\linewidth}\raggedright
Fold 2 MSE
\end{minipage} & \begin{minipage}[b]{\linewidth}\raggedright
Fold 3 MSE
\end{minipage} & \begin{minipage}[b]{\linewidth}\raggedright
Fold 4 MSE
\end{minipage} & \begin{minipage}[b]{\linewidth}\raggedright
Fold 5 MSE
\end{minipage} & \begin{minipage}[b]{\linewidth}\raggedright
Mean MSE
\end{minipage} & \begin{minipage}[b]{\linewidth}\raggedright
SD MSE
\end{minipage} \\
\midrule\noalign{}
\endfirsthead
\toprule\noalign{}
\begin{minipage}[b]{\linewidth}\raggedright
Experiment
\end{minipage} & \begin{minipage}[b]{\linewidth}\raggedright
\(n\) Features
\end{minipage} & \begin{minipage}[b]{\linewidth}\raggedright
Fold 1 MSE
\end{minipage} & \begin{minipage}[b]{\linewidth}\raggedright
Fold 2 MSE
\end{minipage} & \begin{minipage}[b]{\linewidth}\raggedright
Fold 3 MSE
\end{minipage} & \begin{minipage}[b]{\linewidth}\raggedright
Fold 4 MSE
\end{minipage} & \begin{minipage}[b]{\linewidth}\raggedright
Fold 5 MSE
\end{minipage} & \begin{minipage}[b]{\linewidth}\raggedright
Mean MSE
\end{minipage} & \begin{minipage}[b]{\linewidth}\raggedright
SD MSE
\end{minipage} \\
\midrule\noalign{}
\endhead
\bottomrule\noalign{}
\endlastfoot
IG + k-means (most relevant features) & 64 & 6,605 & 5,477 & 5,524 &
5,627 & 5,576 & \textbf{5,762} & 0.425 \\
IG + k-means (most uniformative features) & 22 & 16.795 & 15.872 &
18.692 & 15.498 & 15.537 & \textbf{16.477} & 1.201 \\
Pearson (most corrlating features) & 64 & 13,921 & 15,743 & 11,82 &
13,731 & 9,916 & \textbf{13.026} & 1.990 \\
Lasso Regression & 64 & 7.338 & 8.775 & 8.907 & 8.345 & 8.849 &
\textbf{8.443} & 0.587 \\
KernelShap (most relevant features) & 64 & 7.113 & 8.657 & 7.238 & 9.268
& 6.287 & \textbf{7.824} & 1.089 \\
\end{longtable}

\subsection{Feature Importance
Discussion}\label{feature-importance-discussion}

\begin{figure}[H]

{\centering \includegraphics[width=1\textwidth,height=\textheight]{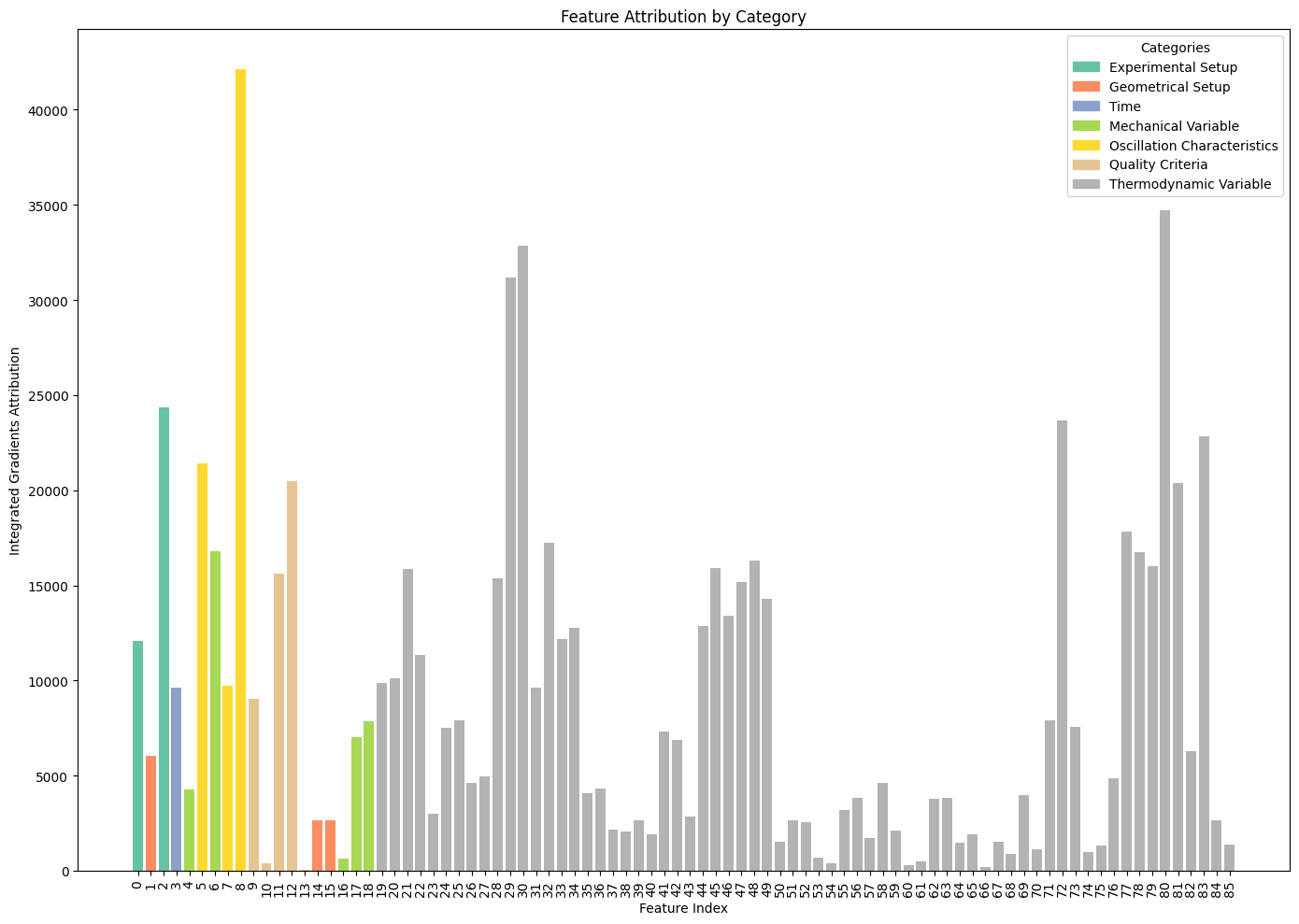}

}

\caption{Absolute feature attribution values based on Integrated
Gradients for the subcategories of the data. The attribution values are
accumulated over the entire dataset, with different colors representing
different feature categories. \label{f_imp2}}

\end{figure}

While our primary goal is to use feature attribution for feature
selection, these methods also provide valuable insights into the
decision-making process of the network, potentially revealing previously
unknown aspects of the underlying physical process. Figure \ref{f_imp2}
illustrates the feature importance across various feature categories as
outlined in Section~\ref{sec-data}. It is evident from the figure that
IG identifies the oscillation mode (index 8) from the ``Oscillation
Characteristics'' group as the most significant variable for predicting
the updated oscillation amplitude. From an engineering perspective, this
is entirely logical, as the mode (the natural frequency) is a key
characteristic of the oscillation. Within the same group, another
crucial feature is the excitation order (index 5). From the category of
thermodynamic variables, several features stand out as influential in
the neural network's prediction process. These include two temperatures
measured on the drive train (indices 80 and 83), the mass flow of air in
front of the compressor (index 30), the volume flow of air in front of
the compressor (index 29), and the pressure of the lubricating oil
(index 72). The high importance of the temperature values measured on
the drivetrain is particularly interesting. This could be explained by a
delayed correlation with the rotation speed. However, it is unexpected
that the measured temperatures appear to be more important for the
neural network's decision-making than the measured rotation speed (index
18) itself. Additionally, the feature that describes the name and
position of the strain gauges (index 2) from the ``Experimental Setup''
category provides valuable information for an accurate prediction of the
amplitude. Moreover, one feature from the ``Quality Criteria'' group
(index 12) has also proven to be particularly important for the
predictions of the neural network.

\section{Conclusion}\label{conclusion}

Even though neural networks possess structural feature engineering
capabilities, this work demonstrates that feature attribution serves as
an effective tool for explainable feature selection. This approach
enhances model performance and stability by identifying informative
features and filtering out noise and disturbances in the data. By
combining feature attribution with clustering techniques, we present a
data-driven method for categorizing features based on their importance,
thereby eliminating the need for arbitrary thresholds in feature
selection.

Our approach serves as an embedded feature selection method for neural
networks, utilizing the model's gradient information to evaluate feature
importance directly from its decision-making processes. This strategy
enhances predictive accuracy and robustness by eliminating noise in the
input, which in turn reduces the standard deviation of the error.
Additionally, it offers valuable insights into the model's reasoning,
promoting the development of more transparent and interpretable AI
systems.

While the primary advantage of our method is the transparency it brings
to feature selection, making AI systems more explainable, our initial
comparative experiments demonstrate that this approach also outperforms
classical feature selection methods. However, it is important to note
that these comparisons were made against standard methods and were based
on the best results of our approach. In future studies, we plan to build
on the findings of this work by conducting more comprehensive
comparative experiments with advanced feature selection methods.
Additionally, we aim to enrich our method by incorporating other feature
attribution techniques.

Despite the promising results, it should not be ignored that the quality
of XAI methods depends on the initial prediction quality of the network.
There may be use cases where the prediction with the initial feature set
is insufficient, and conventional feature selection methods need to be
applied before feature attribution methods. The relatively good results
from the initial evaluation with all features in our study can be
attributed to the feature engineering capabilities of neural networks.

Another interesting finding of this study is the potential to predict
oscillation amplitudes using AI methods. Even though this was not the
main goal of our work, this method seems to offer significant added
value for the development process of turbomachinery. Future research can
therefore focus on fine-tuning the prediction performance by
investigating other machine learning models, more complex network
structures and further hyperparameter tuning.

\section*{Appendix}\label{appendix}
\addcontentsline{toc}{section}{Appendix}

\includegraphics{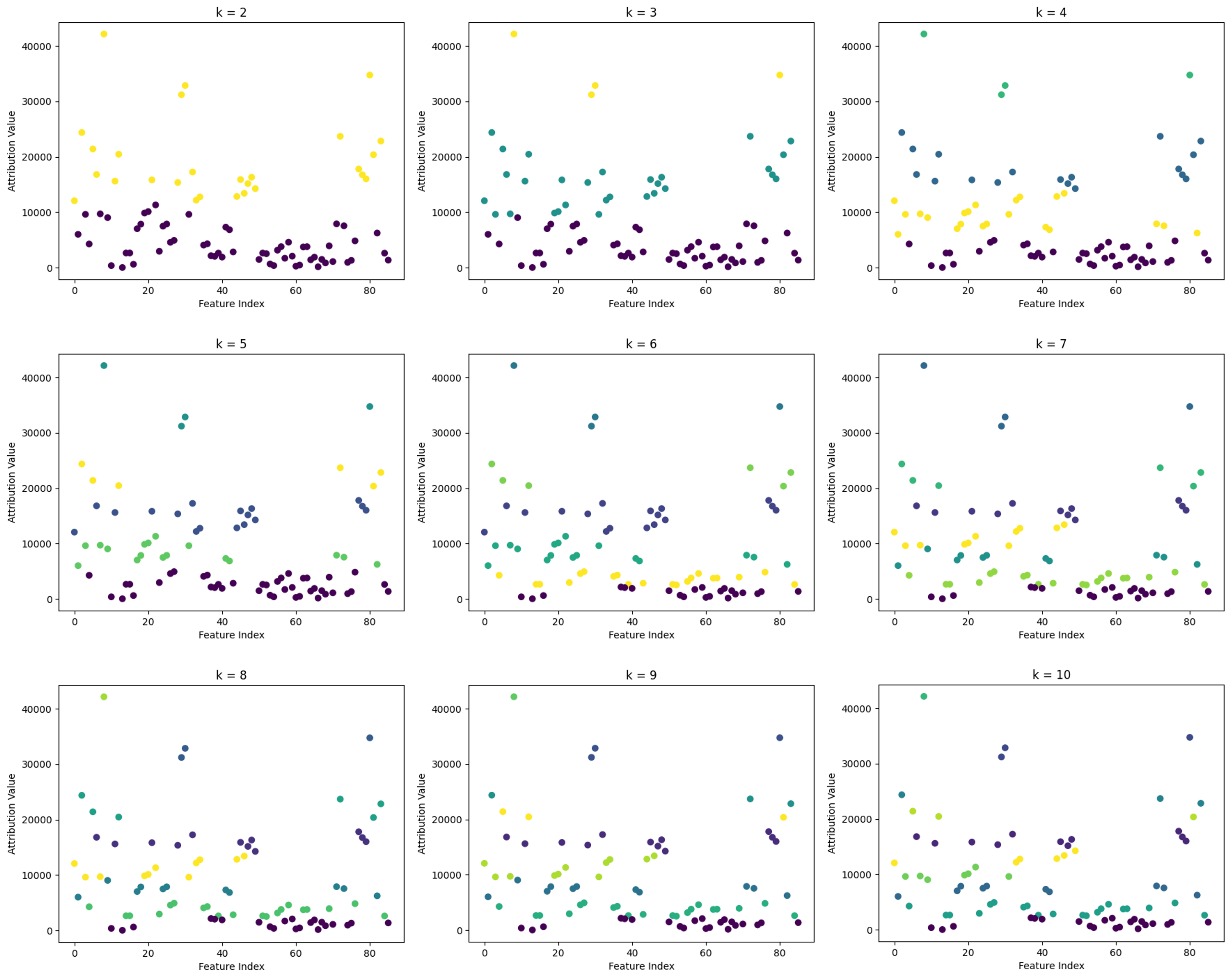} K-Means clustering for the
Integrated Gradient based feature attribution values. The attribution
values are calculated for the tuned neural network on all 86 input
features. The K-Means analyses are performed with different k-values
(k=3 - k=10). All dots of one colour represent a cluster. In all
subgraphs, the violet cluster is related to the least important
features, which are removed for the subsequent evaluation.

\renewcommand\refname{References}
\bibliography{bibliography.bib}

\end{document}